\documentclass[electronic]{vgtc}             
\graphicspath{{figures/}{pictures/}{images/}{./}} 

\usepackage{times}

\usepackage{tabu}                     
\usepackage{booktabs}                  
\usepackage{lipsum}                   
\usepackage{mwe}                       
\usepackage{amsmath}
\usepackage{booktabs}
\usepackage{makecell}
\usepackage{graphicx}

\usepackage{mathptmx}                 
\onlineid{0}
\vgtccategory{Research}
\vgtcinsertpkg

\title{Enhancing Virtual Agents through SLMs and Edge-Computing: An Exploratory Evaluation of Think and Memory Processes}

\author{Aimilios Hadjiliasi\thanks{e-mail: AHadjiliasi1@uclan.ac.uk} 
\and Louis Nisiotis\thanks{e-mail:LNisiotis@uclan.ac.uk} %
}
\affiliation{\scriptsize Department of Computing, Engineering and Mathematics \\ University of Central Lancashire, Cyprus}

\teaser{
  \centering
  \includegraphics[width=\linewidth]{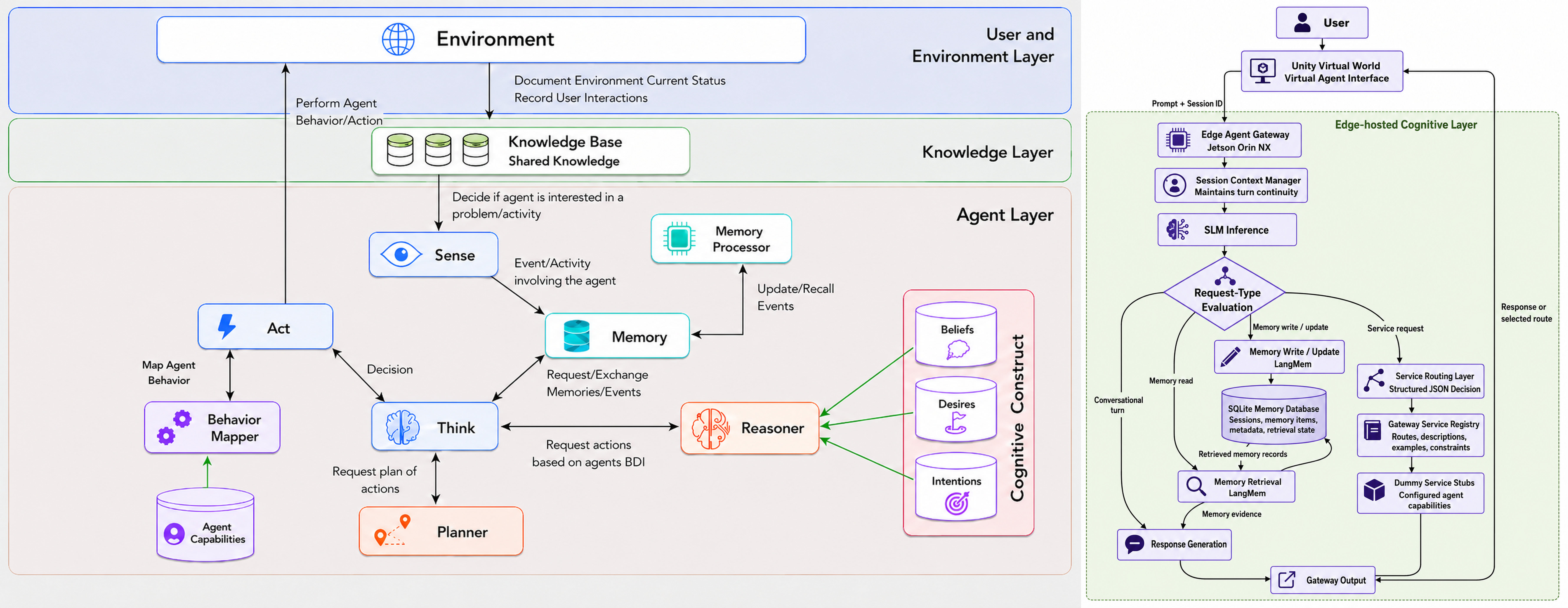}
  \caption{High-level overview of the CEAA architecture (left) and the edge-hosted agent implementation (right).}
  \label{fig:teaser}
}

\abstract{
Embodied intelligent virtual agents are expected to operate as persistent, adaptive, and context-aware entities within complex virtual and Metaverse worlds. 
However, implementing cognitively capable agents in such environments is conceptually and technologically challenging. 
Among a range of blueprints and development approaches, the Cognitive Embodied Agent Architecture (CEAA) has been developed as an implementation-oriented framework for architecting components of perception, memory, reasoning, planning, and embodied action.
Considering the recent advances in edge computing and generative AI language models, this paper explores the use of Small Language Models (SLMs) to support edge-based operation of selected CEAA components, focusing on \textit{Think} and \textit{Memory} as processes central to cognitive orchestration and persistence of virtual agents in interactive virtual worlds. 
An edge-based virtual agent gateway system was developed and evaluated on an NVIDIA Jetson Orin NX using Qwen2.5 models of different sizes, exploring the system's capability to process service requests and handle memory-driven conversations. 
A series of simulation experiments evaluated routing accuracy, memory-read performance, and latency, demonstrating an SLM-driven prototype agent system that partially implements selected CEAA processes to support the development of embodied agents whose cognitive “brain” can operate efficiently and contextually for interactive experiences in immersive virtual worlds.
} 


\begin{document}

\firstsection{Introduction}
\maketitle
Recent advancements in Artificial Intelligence (AI), eXtended Reality (XR), and Metaverse technologies are reshaping how users interact with complex interactive virtual environments and with each other.
Users are no longer limited to static content and predefined scenarios, but share interactive virtual worlds and spaces with other users through avatars, digital twins, and smart interactive objects \cite{dwivedi2022metaverse}. 
In such worlds, virtual agents are intelligent entities situated in virtual environments and represented through embodied forms, and can support communication, guidance, assistance, training, education, and social interaction, operating as persistent, adaptive, and context-aware entities capable of responding to users and environmental changes in real-time \cite{griol2019developing}.
However, implementing cognitively capable embodied agents in real-time virtual environments remains challenging, as many still rely on scripted behaviors, rule-based flows, and symbolic game AI techniques.
Although effective for predefined behaviors, these approaches often limit the agent’s ability to adapt to dynamic user input and maintain continuity across interactions. 
As a result, agents may appear embodied in the visual sense, but remain limited in cognitive and interactive capabilities.
To support the development of intelligent interactive agents, the \textit{Cognitive Embodied Agent Architecture (CEAA)} has been proposed as an implementation-oriented framework (see Figure ~\ref{fig:teaser} (left)) \cite{hadjiliasiCEAA2026}, to support the development of intelligent interactive agents.
CEAA builds on existing frameworks and defines the architectural components required for perception, memory, reasoning, planning, behavior mapping, and embodied action, providing a modular foundation for moving towards cognitively capable virtual agents. 
Building on that, this paper explores some of the recent advancements in generative AI, particularly Small Language Models (SLMs), as a cognition mechanism to support the edge-based operation of selected CEAA components for intelligent virtual agents.  
Edge-based processing is essential for responsive in-world interaction, enabling embodied agents to interpret input, access memory, infer intent, and coordinate responses with minimal latency. 
SLMs are well suited for this role because they offer efficient inference with lower computational demands and can operate closer to the user.
Specifically, this exploratory study examines the use of SLMs to implement two key CEAA processes driving intelligent agent behaviour: \textit{Think} and \textit{Memory}, which are central to cognitive orchestration and persistence.
To explore this, an edge-based virtual agent gateway was developed to process user prompts through locally deployed SLMs. 
The system supports memory handling and service routing, allowing agents to store and retrieve context and select suitable backend services for conversational support and generative tasks, such as handling user requests for 3D models, textures, motion, sound, and text-to-speech services during conversation.
Accordingly, this exploratory study is guided by the following research question: 
\textbf{RQ: To what extent can SLMs partially operationalize selected aspects of CEAA’s \textit{Think} and \textit{Memory} components through service routing and structured memory handling for an edge-based virtual-world agent?}
To ascertain this, the study evaluates Qwen2.5 small models of different sizes in an edge-computing setup, examining the relationship between model size, routing accuracy, memory performance, and latency under resource-constrained conditions. 
As such, this paper contributes by: i) partially implementing and evaluating selected aspects of CEAA’s \textit{Think} and \textit{Memory} components through service routing and structured factual memory handling; ii) examining the feasibility of these mechanisms under edge-computing conditions; and iii) providing early evidence on how selected backend processes of embodied agents can operate locally and contextually.

\section{Background and Context}
\subsection{Virtual Agents} 
Virtual agents are computer-controlled entities in virtual environments that interact with users, other agents, and the surrounding space. 
Based on agent theory, they are considered intelligent when they act autonomously, perceive their environment, pursue goals, and demonstrate reactive, proactive, and social behavior \cite{wooldridge1995intelligent}.
In virtual worlds, this definition is extended through visual, embodied, and behavioural representation \cite{hadjiliasi2025use}. 
Accordingly, embodied agents should communicate through speech, gaze, gestures, facial expressions, locomotion, and other multimodal behaviours.
To support this, agent behaviour should be driven by perception, reasoning, memory, dialogue, planning, and intelligent decision-making \cite{wooldridge1995intelligent,funge1999cognitive}. 
Early attempts on virtual agents focused on animated entities to guide users, demonstrate tasks and content, and support face-to-face interaction, a role that remains relevant in Metaverse-type virtual world environments \cite{hadjiliasi2025use}. 
Over time, agents have evolved from scripted and rule-based characters toward more cognitively capable systems, supported by architectures such as the BDI \cite{rao1995bdi} and SOAR \cite{laird1987soar} multimodal frameworks \cite{kopp2006towards,hartholt2013all}, and, more recently, Large Language Models (LLMs) \cite{park2023generative}.
Embodiment is key in virtual environments because it changes how users perceive and interact with agents, as it can make interaction feel closer to communication with a social partner, and empirical work has shown that well-designed agents can enhance user engagement \cite{yang2025embodied}.
Therefore, agent development and evaluation should consider user experience as well as functionalities. 
In this paper, we are focusing on their functionalities.
Technically, agents can be evaluated through task success, conversational ability, dialogue accuracy, memory recall, latency, robustness, and error rates. 
Despite progress, key challenges remain, including real-time integration of perception, memory, reasoning, and embodied action, natural behavior, reliable context handling, reduced hallucinations, safe outputs, and user privacy.
These challenges motivate further research into architectures and implementation mechanisms to enable the development of persistent, adaptive, context-aware, and robust virtual agents, safe to deploy in virtual worlds and complex computing systems.

\subsection{Memory and Persistence in Virtual Agents}
Memory and information persistence have become important research areas in agent systems, where memory is often divided into distinct types. 
From a cognitive science perspective, this includes \textit{working memory}, which supports immediate reasoning, and \textit{long-term memory}, which stores information beyond the current interaction \cite{baddeleyA1992WorkingMemory, tulving1972episodic}.
Within the scope of language model driven agents, MIRIX memory system for instance extends this distinction by proposing six specialized memory components: \textit{Core Memory} that stores persistent high-priority information about the agent and user; \textit{Episodic Memory} that stores time-stamped events and interactions; \textit{Semantic Memory} that stores concepts, entities, and relationships; \textit{Procedural Memory} that stores task procedures and workflows; \textit{Resource Memory} that stores documents, files, and media, and last; \textit{Knowledge Vault} that preserves sensitive information such as credentials, addresses, or contact details \cite{wang2025mirix}. 
This distinction is important because persistent virtual agents require more than conversation history. 
To operate, they require structured mechanisms for storing, retrieving, updating, and protecting different kinds of user, task, and environment knowledge \cite{wang2025mirix}. 
Agent memory also enables agents to encode, store, retrieve, and use information from previous interactions, supporting continuity, identity, user preferences, task history, and persistence across sessions \cite{wang2025mirix,park2023generative,maharana2024evaluating}. 
In virtual worlds, this allows agents to remember users, objects, locations, actions, relationships, and ongoing tasks.
Generative-agent research shows that memory, reflection, and planning can improve behavioral believability in simulated social environments. 
However, long-term conversational evaluations reveal persistent limitations in temporal reasoning, causal consistency, and multi-session recall \cite{park2023generative, maharana2024evaluating}. 
Virtual agents therefore require robust memory mechanisms to reliably store, retrieve, update, and reason over contextual information across interactions.

\subsection{SLMs and Edge-Based Agentic AI}
SLMs are lightweight transformer-based models, typically containing hundreds of millions to several billion parameters. Compared with larger LLMs, they support language understanding and generation with lower computational, memory, latency, and power requirements \cite{lu2024small}. These characteristics make them suitable for efficient agentic tasks, including structured output generation, memory-operation detection, summarisation, dialogue handling, information extraction, task decomposition, intent recognition, request classification, and service routing \cite{lu2024small,belcak2025small,nisiotis2026prompt}.

In agent systems, particularly virtual worlds, SLMs can operate beyond conversational response generation as lightweight cognitive controllers. They can interpret user input, classify intent, determine whether memory should be accessed or updated, and select the appropriate processing route. This is especially relevant in complex virtual environments, where requests may require different backend capabilities, such as conversation, memory retrieval, 3D generation, or other specialized functions. In such cases, the agent must orchestrate tools, services, and specialized models rather than only generate dialogue \cite{nisiotis2026prompt}.

However, SLMs remain limited in deep contextual reasoning, long-term planning, broad factual coverage, complex abstraction, and the robust handling of ambiguous or open-ended requests. Their smaller size may also reduce consistency across extended interactions, knowledge integration, and creative output quality. Larger LLMs are therefore more appropriate for complex open-ended reasoning, multi-step planning, broad knowledge synthesis, high-quality creative generation, and multimodal understanding \cite{belcak2025small,yao2022react}.

On the other hand, edge-based agentic AI refers to agent systems where inference, memory handling, decision logic, tool use, or orchestration are close to the user and environment rather than on the cloud \cite{zhou2019edge}. 
In virtual environments, this supports responsive, privacy-sensitive, and network-resilient interaction. 
Architecturally, SLM-based edge agents often combine a local language model, memory or retrieval module, tool-calling interface, task router, and optional cloud fallback \cite{zhou2019edge,nisiotis2026prompt}.
Through tool use and function calling, the model can move beyond text generation by producing structured API calls, invoking external services, receiving observations, and using returned results in subsequent reasoning and decision-making. 
This allows agentic systems to delegate specialised tasks to appropriate tools, services, or models rather than attempting to compute all functions internally.
To explore this direction, this paper develops an edge-based virtual agent system, focusing on service routing and memory capabilities to examines whether SLMs can support these backend cognitive processes locally, enabling virtual agents to interpret user requests, preserve context, and coordinate access to generative services for responsive in-world interaction.

\subsection{CEAA: Cognitive Embodied Agent Architecture}
Developing intelligent virtual agents requires architectures that can organize perception, memory, reasoning, decision-making, and embodied action. 
Several approaches have been proposed for this purpose, including rational agent architectures such as BDI \cite{rao1995bdi}, cognitive architectures such as SOAR \cite{laird1987soar}, embodied conversational-agent frameworks such as SAIBA and Greta \cite{kopp2006towards,niewiadomski2009greta}, and virtual-human development toolkits such as the ICT Virtual Human Toolkit \cite{hartholt2013all}. 
These approaches provide important foundations for agent reasoning, multimodal behaviour, and interactive virtual humans. 
However, integrating cognitive processes with real-time embodied execution in interactive virtual worlds remains challenging, especially when agents must operate persistently, respond to dynamic user input, access memory, and coordinate actions or services during runtime.

CEAA \cite{hadjiliasiCEAA2026} is an implementation-oriented architecture proposed for developing cognitive embodied intelligent virtual agents that operate in real-time interactive virtual environments (see Figure ~\ref{fig:teaser} (left)). 
It was introduced to bridge the gap between low-level reactive implementations, such as finite-state machines and symbolic/game-AI techniques, and high-level cognitive architectures that provide rich reasoning models but are often difficult to integrate into real-time 3D environments. 
As such, CEAA connects cognitive reasoning with embodied execution by providing a reusable framework for implementing the “brain” of virtual agents in complex interactive systems and Metaverse applications.

CEAA consists of three main layers: the \textit{User and Environment} layer, the \textit{Knowledge} layer, and the \textit{Agent} layer. 
The \textit{User and Environment} layer represents the virtual world, including users, agents, objects, and system-level events. 
The \textit{Knowledge} layer maintains a structured representation of the environment through a shared blackboard-oriented knowledge base, where events and state changes are recorded and made available to the agent. 
The \textit{Agent} layer contains the cognitive and behavioural components that allow the agent to sense events, access memory, think, reason, plan, map decisions to embodied behaviour, and act within the virtual environment. 
In this way, CEAA separates environmental dynamics, shared knowledge, cognitive processing, and embodied action into modular components that can be implemented in development environments.

Within CEAA, the \textit{Memory} and \textit{Think} components are central to persistent and adaptive agent behaviour. 
\textit{Memory} stores and organises past experiences, including episodic information, semantic knowledge, user-related information, and prior actions, enabling the agent to retrieve relevant events and adapt its behaviour based on previous interactions. 
\textit{Think} acts as the agent’s central cognitive coordinator by integrating information from memory and coordinating with modules such as the Reasoner and Planner to determine the most appropriate action. 
Instead of relying on a single reasoning algorithm, it orchestrates the flow of information between perception, memory, reasoning, planning, and action selection.
Building on this foundation, this paper aims to explore and enhance the practical operation of these components through edge-based computing and SLMs to support persistent, adaptive, and context-aware behaviour under real-time and resource-constrained conditions.

\section{Methodology}
This paper investigates the use of SLMs to support the edge-based operation of CEAA’s \textit{Think} and \textit{Memory} components, for the development of interactive conversational virtual agents capable of holding memory-based conversations and identifying user requests for external services.
Our previous work introduced an SLM-based Agent Orchestration Gateway for routing virtual-world requests to heterogeneous AI services \cite{nisiotis2026prompt}. 
The present study extends this architecture by integrating contextual memory and systematically comparing three general-purpose SLM sizes across broader routing and memory tasks.
The \textit{Think} component is partially implemented through service routing as a form of cognitive orchestration, and \textit{Memory} is examined through structured write--read interaction handling.
The evaluation focuses on these components as practical mechanisms through which an embodied virtual agent can interpret user input, determine the appropriate processing pathway, and store, retrieve, or update contextual information across interactions. 
To support this, an edge-based agent gateway was developed, which receives user prompts, processes them using Qwen2.5 models, and returns either a structured routing decision or a memory-oriented response.
Three model sizes (0.5B, 1.5B, and 3.0B parameters) were evaluated to analyse the relationship between model size, routing accuracy, memory performance, structured-output reliability, and interaction latency under resource-constrained conditions.  
To explore this, an edge-based agent system is integrated with a 3D virtual avatar, where the SLM acts as the agent’s cognitive “brain”, with regard to CEAA's "think" and "memory" processes. 
The system enables real-time input interpretation, service routing, memory access and updates, and coordination with backend generative services.
This setup allows the evaluation of SLM-driven \textit{Think} and \textit{Memory} processes within a situated and embodied context, where decisions made by the model directly influence in-world agent behaviour and interaction flow.

\subsection{System Configuration and Apparatus}
The experimental apparatus extended the CEAA-based \textit{InterwovenXR} virtual-world testbed developed through the authors’ ongoing work on intelligent virtual environments. 
The testbed has supported virtual museums, robotic digital twins, and other embodied-agent scenarios, providing a reusable Unity-based environment where agents interact with users, virtual objects, and system events \cite{nisiotis2023interwoven, hadjiliasi2024comparative}. 
Considering the system used for experimentation, Unity provided the embodied-agent interface through which users submitted natural-language requests and received responses. Language processing, memory management, route selection, and service dispatch were handled externally by an edge-hosted gateway, keeping the virtual-world client lightweight. The gateway and local model server ran on an NVIDIA Jetson Orin NX 8GB, representing resource-constrained edge hardware for virtual-world agents.

Each HTTP request contained a user prompt and session identifier. Qwen2.5 first classified the interaction as conversation, memory-read, memory-write, service request, or ambiguous. This classification determined whether the gateway generated a conversational response, accessed stored context, or selected a configured service route. For service requests, the SLM returned a structured JSON decision containing the selected route, interpreted intent, confidence estimate, and rationale.

The memory subsystem integrated Qwen2.5, LangMem\footnote{https://langchain-ai.github.io/langmem/}, and SQLite. Regardless of the predicted interaction label, every session turn was recorded in SQLite and observed by LangMem, which used the same Qwen2.5 model to extract concise, durable facts. These facts were stored as content-keyed items with session identifiers, timestamps, and metadata. Exact duplicates updated the existing item, whereas non-identical corrections were retained as newer records without automatically deleting earlier information.

For memory-read requests, the gateway deterministically retrieved relevant facts and recent session turns based on token relevance and recency. These records were supplied directly to Qwen2.5 for response generation without additional LangMem filtering. The evaluated memory accuracy therefore reflects the complete classification–extraction–persistence–retrieval–generation pipeline.

\begin{figure}[!hb]
    \centering
    \includegraphics[width=\columnwidth]{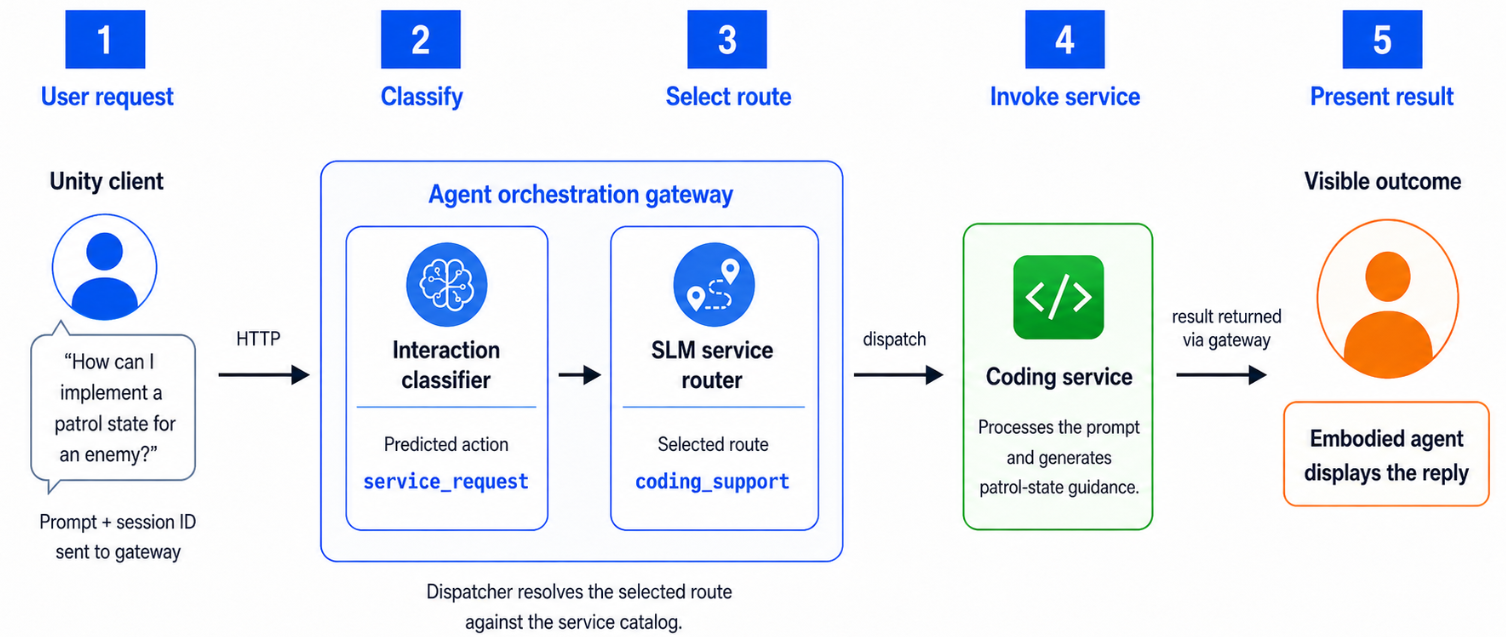}
    \caption{Illustrative gateway flow from user request to coding-service response presented by the embodied agent.}
    \label{fig:interaction-flow}
\end{figure}

For the controlled routing evaluation, ten configured routes represented common virtual-world services (Table \ref{tab:routing_per_route_f1_comparison}).
Each route was defined by a name, description, examples, and routing constraints. 
The routes acted as service stubs: the experiment assessed whether the SLM selected the intended target without executing the downstream generative service. 
Figure \ref{fig:interaction-flow} illustrates the complete deployment flow, including service invocation and presentation of the returned output, whereas the reported routing experiment ended after route selection.

All experiments used the instruction-tuned \texttt{Qwen / Qwen2.5 - {0.5B, 1.5B, 3B} - Instruct - GGUF} checkpoints with Q4\_K\_M quantisation. 
The models were served locally through \texttt{llama.cpp} with a 4,096-token context. 
All calls were limited to 420 generated tokens. 
Top-$p$ and top-$k$ followed the server defaults, and generation terminated at the model end token or token limit. 
Complete system and routing prompts, service descriptions, and memory instructions are provided as supplementary materials\footnote{https://github.com/AimiliosHadjiliasis/CEAA/tree/main/XRAG2026}.

\subsection{Routing Evaluation}
The routing evaluation assessed each SLM’s ability to support the CEAA \textit{Think} process by selecting the appropriate route for user requests. An automated script submitted a balanced dataset of 1,000 prompts to the gateway in route-only mode, with 100 prompts for each of ten predefined routes. Prompt creation was LLM-assisted, followed by author review and revision to ensure relevance to the collaborative game-development scenario and assignment of one ground-truth route per prompt. Downstream services were not invoked, isolating routing behaviour from service execution.

The routes represented general conversation, coding support, gameplay mechanics, game AI guidance, 3D generation, image-to-3D generation, motion generation, texture generation, sound generation, and text-to-speech. For each prompt, the script recorded the predicted and expected routes, correctness, confidence score, JSON validity, and end-to-end routing latency.

Performance was measured using overall accuracy and macro-averaged precision, recall, and F1-score, with macro-averaging ensuring equal treatment of frequent and infrequent routes. Robustness was assessed through invalid-output and timeout rates. Latency was measured in milliseconds and summarised using mean, median, and P95 values. Routing latency included request submission, prompt processing, model inference and decoding, output parsing, and return of the routing decision, while excluding downstream service execution. For memory interactions, the aggregate latency additionally included the applicable extraction, persistence, retrieval, database-access, context-construction, and response-generation operations.

\subsection{Memory Evaluation}
The memory evaluation assessed each SLM’s ability to support the CEAA \textit{Memory} process through structured write–read interactions. It simulated a collaborative virtual-world game-development session in which users progressively introduced information for the agent to store, retrieve, and update.

End-to-end recall was evaluated using 250 prompts: 125 fact-introduction turns and 125 paired memory questions. \textit{End-to-end read accuracy} measured the proportion of questions answered correctly using retrieved memories and session context, with correct reads also reported as a raw count out of 125. \textit{Memory-write label recall} measured the proportion of fact-introduction prompts assigned the memory-write action label. Content persistence was not evaluated because all turns were recorded and independently observed for fact extraction. \textit{Service leakage} counted memory prompts incorrectly routed to external services. Memory latency was reported using mean, median, and P95 end-to-end response times, covering the applicable extraction, persistence, retrieval, database access, context construction, and response generation.

Each write prompt introduced one fact concerning design decisions, responsibilities, preferences, character identities and behaviours, system behaviours, associative links, task updates, or corrections. The paired read prompt queried the same fact using different wording, testing retrieval rather than surface repetition.

Prompt pairs were grouped into design facts, ownership and responsibilities, team interests, task updates, character identity, character behaviour, associative links, state behaviour, and corrections or overrides. These categories covered both basic recall and more demanding behaviours, including updating outdated information, distinguishing related memories, and preserving entity–attribute relationships.

Each pair included a predefined expected fact and was reviewed for ambiguity, scenario relevance, and alignment with the intended memory behaviour. In June 2026, ChatGPT 5.5 High Reasoning evaluated all 250 memory-read responses using a fixed prompt that checked for the expected fact, contradictions, outdated information, and unsupported content. The authors then manually reviewed all outputs using a model-blinded approach.

Responses were correct when they retrieved the requested information without substituting stale, adjacent, or unrelated content. They were incorrect when they omitted the expected fact, returned outdated information, confused related entities, acknowledged a memory action instead of recalling information, or introduced unsupported claims. Ambiguous cases, including incorrect related retrievals, unapplied corrections, and recall requests misclassified as write actions, were manually inspected by the authoring team.

\section{Results}
\subsection{Routing Evaluation}
The routing results (Table \ref{tab:routing_model_comparison}) show a clear performance increase with model size. 
The 0.5B model achieved 29.4\% accuracy and 28.9\% macro-F1, indicating unreliable routing despite producing valid structured outputs. 
Its errors were dominated by route collapse, frequently misclassifying prompts as \texttt{image\_to\_3d\_generation} or \texttt{conversational\_generator}, suggesting limited capability in separating closely related services.
The 1.5B model significantly improved performance, reaching 85.4\% accuracy and 84.3\% macro-F1, and was able to separate most routes reliably. 
The 3.0B model achieved the highest performance (87.7\% accuracy, 87.8\% macro-F1) with no invalid outputs, improving particularly on challenging generation-related routes.
This improvement comes at a latency cost, however, with mean latency increasing from 3349 ms (1.5B) to 5067 ms (3.0B). 
Table \ref{tab:routing_per_route_f1_comparison} shows the results of the different SLM variants to achieve per-route identification and reveals substantial variation across routes. 
The 0.5B model performed poorly on most categories, particularly \texttt{generation\_3d} and \texttt{gameplay\_mechanics}, although it performed comparatively better on \texttt{text\_to\_speech} and \texttt{sound\_generation}. 
The 1.5B model produced strong improvements across nearly all routes, but
\texttt{image\_to\_3d\_generation} remained challenging.
The 3.0B model achieved the highest score on the majority of the, particularly improving generation-related services, although the 1.5B model remained stronger for \texttt{game\_ai\_guidance}, \texttt{gameplay\_mechanics}, and \texttt{conversational\_generator}.

\begin{table}[!hb]
\caption{Routing-only evaluation results across model sizes.}
\label{tab:routing_model_comparison}
\centering
\scriptsize
\setlength{\tabcolsep}{2pt}
\renewcommand{\arraystretch}{0.95}
\begin{tabular}{@{}lccccrrr@{}}
\toprule
\rotatebox{90}{\makecell[l]{Model}} &
\rotatebox{90}{\makecell[l]{Accuracy}} &
\rotatebox{90}{\makecell[l]{Macro\\precision}} &
\rotatebox{90}{\makecell[l]{Macro\\recall}} &
\rotatebox{90}{\makecell[l]{Macro\\F1}} &
\rotatebox{90}{\makecell[l]{Mean\\latency\\(ms)}} &
\rotatebox{90}{\makecell[l]{Median\\latency\\(ms)}} &
\rotatebox{90}{\makecell[l]{P95\\latency\\(ms)}} \\
\midrule
0.5B & 29.4\% & 58.4\% & 29.4\% & 28.9\% & 1504 & 1623 & 1810 \\
1.5B & 85.4\% & 88.9\% & 85.4\% & 84.3\% & 3349 & 3519 & 3923 \\
3.0B & 87.7\% & 90.8\% & 87.7\% & 87.8\% & 5067 & 5367 & 6159 \\
\bottomrule
\end{tabular}
\end{table}

\begin{table}[!hb]
  \caption{Per-route success comparison across Qwen2.5 model sizes.}
  \label{tab:routing_per_route_f1_comparison}
  \centering
  \scriptsize
  \begin{tabular}{lccc}
    \toprule
    Route & 0.5B & 1.5B & 3.0B \\
    \midrule
    coding\_support & 13.5\% & 90.4\% & 96.4\% \\
    game\_ai\_guidance & 47.3\% & 78.6\% & 76.5\% \\
    gameplay\_mechanics & 5.8\% & 85.7\% & 78.6\% \\
    generation\_3d & 0.0\% & 75.5\% & 80.9\% \\
    image\_to\_3d\_generation & 27.0\% & 45.0\% & 74.1\% \\
    motion\_generation & 24.6\% & 97.6\% & 99.5\% \\
    conversational\_generator & 10.2\% & 85.6\% & 72.9\% \\
    sound\_generation & 65.5\% & 93.4\% & 100.0\% \\
    text\_to\_speech & 78.8\% & 95.3\% & 99.5\% \\
    texture\_generation & 16.5\% & 95.7\% & 99.0\% \\
    \bottomrule
  \end{tabular}
\end{table}

\subsection{Memory Evaluation}
The memory evaluation followed the same comparative structure across the three model sizes (see Table \ref{tab:v20_2_memory_comparison}).
The 0.5B model achieved 72.8\% memory-read accuracy (91/125), but while it performed well on simple facts, it struggled with correction handling, often retrieving the correct entity but the wrong memory facet, indicating limited fine-grained memory selection.
The 1.5B model improved to 78.4\% accuracy (98/125), showing better performance on factual recall, associative links, and corrections. 
However, it remained weak on task updates and state behaviour, suggesting difficulty in replacing or disambiguating closely related information.
The 3.0B model achieved the highest performance at 93.6\% (117/125), handling facts, state behaviour, and corrections reliably. 
Some errors were mainly due to misclassification of recall prompts as memory-write actions, indicating that improved memory fidelity address mostly several issues of classification.
This performance gain comes with a latency trade-off however, with the 3.0B model achieving the highest accuracy but with substantial latency (mean 9693 ms, P95 14531 ms). 
The 0.5B model was significantly faster but unsuitable due to poor correction handling and low write-action recognition. 
The 1.5B model provided a latency compromise but showed the weakest action consistency, with many memory interactions treated as generic responses rather than explicit memory operations.

\begin{table}[tb]
\caption{Comparative memory evaluation results.}
\label{tab:v20_2_memory_comparison}
\centering
\scriptsize
\setlength{\tabcolsep}{2pt}
\renewcommand{\arraystretch}{0.95}
\begin{tabular}{@{}lccccrrr@{}}
\toprule
\rotatebox{90}{\makecell[l]{Model}} &
\rotatebox{90}{\makecell[l]{End-to-end\\read accuracy}} &
\rotatebox{90}{\makecell[l]{Correct\\reads}} &
\rotatebox{90}{\makecell[l]{Memory-write\\label recall}} &
\rotatebox{90}{\makecell[l]{Service\\leakage}} &
\rotatebox{90}{\makecell[l]{Mean\\latency\\(ms)}} &
\rotatebox{90}{\makecell[l]{Median\\latency\\(ms)}} &
\rotatebox{90}{\makecell[l]{P95\\latency\\(ms)}} \\
\midrule
0.5B & 72.8\% & 91/125  & 1.6\%  & 1/250 & 2025 & 1845 & 2385  \\
1.5B & 78.4\% & 98/125  & 5.6\%  & 7/250 & 3794 & 3582 & 5416  \\
3.0B & 93.6\% & 117/125 & 63.2\% & 8/250 & 9693 & 9175 & 14531 \\
\bottomrule
\end{tabular}
\end{table}

\section{Discussion}
The results show that SLMs possess inference capabilities that can support selected CEAA \textit{Think} and \textit{Memory} processes. For \textit{Think}, the 1.5B and 3.0B models effectively supported intent classification and routing across conversational, memory-related, and external generative services hosted on different servers or hardware. This aligns with agentic AI and tool-use research, in which language models select actions within structured action spaces \cite{yao2022react,schick2023toolformer}. However, ambiguities remained between semantically similar services, indicating that model suitability depends on the target function, model size, and latency requirements.

For \textit{Memory}, Qwen2.5-3.0B achieved the highest read accuracy and reliably handled factual recall, state behaviour, and corrections, consistent with prior research on memory, reflection, and planning in believable agents \cite{park2023generative,maharana2024evaluating}. Nevertheless, some recall requests were misclassified as memory-write actions, showing that memory fidelity and memory-action classification remain distinct challenges.

The findings also reveal a clear accuracy–responsiveness trade-off. The 0.5B model was the fastest but unreliable for routing and correction handling; the 1.5B model provided the best routing balance; and the 3.0B model achieved the strongest memory performance but incurred substantial latency. CEAA-based agents may therefore benefit from modular or hybrid configurations that use smaller models for low-latency orchestration and larger models for memory-intensive or semantically complex tasks.

Building on these findings, the study addresses the research question by demonstrating how SLMs can support selected CEAA processes and locally implement aspects of an agent’s cognitive ``brain'' through service selection and contextual memory handling at the edge. Challenges remain in routing ambiguity, memory-action classification, latency optimisation, and deployment within live interaction loops. The results provide early evidence that edge-based SLMs can support persistent, adaptive, and context-aware virtual agents, while emphasising the importance of careful model selection and task-specific design.

As such, the study makes three main contributions. First, it empirically demonstrates how service routing can partially implement the CEAA \textit{Think} process and how structured write–read interactions can partially implement \textit{Memory} under edge-computing constraints. This does not constitute a complete demonstration of embodied-agent behaviour, as perception, behaviour mapping, embodied action, and live 3D interaction were outside the evaluation scope. Second, it compares Qwen2.5 models of different sizes for local routing and memory handling, showing through accuracy, memory, and latency results that model selection should depend on the target function. Third, it contributes to the broader vision of complex virtual worlds and Metaverse systems by showing how backend cognitive processes for embodied agents can begin to operate locally, responsively, and contextually under real-time, resource-constrained conditions.
\section{Conclusions, Limitations and Future Directions}
This study provides empirical evidence that locally deployed SLMs can support selected aspects of CEAA’s \textit{Think} and \textit{Memory} components within an edge-hosted embodied-agent backend. 
The results demonstrate the feasibility of local service routing and contextual memory handling, while showing that larger models improve reliability at the cost of increased latency, requiring careful task allocation and further optimisation for interactive use.

However, several limitations remain. 
The evaluation was conducted in a controlled test-bed and therefore did not capture complete embodied user–agent interaction. 
It examined only selected CEAA processes, while the controlled prompt sets may not fully represent unpredictable user behaviour. 
Although the LLM-as-a-judge protocol used predefined expected answers and human verification, evaluation bias may remain. 
The findings are also limited to three Qwen2.5 variants and a single Jetson edge configuration, restricting their generalisability to other SLM families and hardware platforms.

The aggregate latency measurements did not isolate prompt preparation, decoding, fact extraction, SQLite access, retrieval, response generation, serialisation, or communication overhead. 
Future profiling should measure each stage separately, report time to first token and generation throughput, and establish acceptable latency thresholds through user studies with embodied agents.

The study was also limited to general-purpose generative SLMs and did not compare keyword routing, embedding similarity, or task-specific fine-tuned SLMs, which showed potential for service routing in our previous work \cite{nisiotis2026prompt}. 
The results therefore demonstrate feasibility and model-size trade-offs rather than the superiority of generative SLMs.

As such, future work should benchmark these alternatives for service selection and memory-action classification, while independently measuring classification accuracy, extraction fidelity, retrieval recall, correction resolution, and answer accuracy. 
Memory evaluation should also extend beyond paired write–read prompts to richer long-term structures, including episodic, semantic, procedural, spatial, and user-preference memory. 
Finally, human-participant evaluations should assess whether edge-based SLMs can support adaptive, persistent, and context-aware embodied agents in realistic virtual-world scenarios.

\bibliographystyle{abbrv}

\bibliography{template}

@article{dwivedi2022metaverse,
  title={Metaverse Beyond the Hype: Multidisciplinary Perspectives on Emerging Challenges, Opportunities, and Agenda for Research, Practice and Policy},
  author={Dwivedi, Yogesh K and Hughes, Laurie and Baabdullah, Abdullah M and Ribeiro-Navarrete, Samuel and Giannakis, Mihalis and Al-Debei, Mutaz M and Dennehy, Denis and Metri, Bhimaraya and Buhalis, Dimitrios and Cheung, Christy MK and others},
  journal={International journal of information management},
  volume={66}, pages={102542},
  year={2022}, publisher={Elsevier}, doi={10.1016/j.ijinfomgt.2022.102542}
}

@inproceedings{nisiotis2026prompt,
  author    = {Nisiotis, Louis and Hadjiliasi, Andreas},
  title     = {From Prompt to Service: An SLM-Based Agent
               Orchestration Gateway for AI-Driven Virtual Worlds},
  booktitle = {Proceedings of the 2026 IEEE 3rd International
               Symposium on Emerging Metaverse (ISEMV)},
  address   = {Cyprus},
  month     = oct,
  year      = {2026},
  note      = {In press}
}

@article{griol2019developing,
  title={Developing Enhanced Conversational Agents for Social Virtual Worlds},
  author={Griol, David and Sanchis, Araceli and Molina, Jos{\'e} Manuel and Callejas, Zoraida},
  journal={Neurocomputing},
  volume={354},
  pages={27--40},
  year={2019},
  publisher={Elsevier}, doi={https://doi.org/10.1016/j.neucom.2018.09.099}
}

@article{wooldridge1995intelligent, 
    title={Intelligent Agents: Theory and Practice}, 
    author={Wooldridge, Michael and Jennings, Nicholas R}, 
    journal={The Knowledge Engineering Review}, 
    volume={10}, 
    number={2}, 
    pages={115--152}, 
    year={1995}, 
    publisher={Cambridge University Press}, doi ={https://doi.org/10.1017/S0269888900008122}
}

@inproceedings{funge1999cognitive,
    title={Cognitive Modeling: Knowledge, Reasoning and Planning for Intelligent Characters},
    author={Funge, John and Tu, Xiaoyuan and Terzopoulos, Demetri},
    booktitle={Proceedings of the 26th annual conference on Computer graphics and interactive techniques},
    pages={29--38},
    year={1999}
}

@inproceedings{hadjiliasi2024comparative,
    title={A Comparative Assessment of Technology Acceptance and Learning Outcomes in Computer-Based versus VR-based Pedagogical Agents},
    author={Hadjiliasi, Aimilios and Nisiotis, Louis and Polycarpou, Irene},
    booktitle={2024 IEEE International Symposium on Mixed and Augmented Reality Adjunct (ISMAR-Adjunct)},
    pages={513--516},
    year={2024}
}

@inproceedings{niewiadomski2009greta,
  author    = {Niewiadomski, Rados{\l}aw and Bevacqua, Elisabetta and Mancini, Maurizio and Pelachaud, Catherine},
  title     = {Greta: An Interactive Expressive {ECA} System},
  booktitle = {Proceedings of the 8th International Conference on Autonomous Agents and Multiagent Systems (AAMAS 2009)},
  volume    = {2},
  pages     = {1399--1400},
  year      = {2009},
  publisher = {International Foundation for Autonomous Agents and Multiagent Systems},
  url       = {https://dl.acm.org/doi/10.5555/1558109.1558314}
}

@inproceedings{hadjiliasi2025use,
  title={On the Use of Virtual Agents in EduVerse: A Survey of Embodied Virtual Agent Types and Future Research Directions in {E}du-Verse Applications},
  author={Hadjiliasi, Aimilios and Nisiotis, Louis and Polycarpou, Irene},
  booktitle={2025 IEEE International Symposium on Emerging Metaverse (ISEMV)},
  pages={129--138},
  year={2025},
  organization={IEEE}, doi={10.1109/ISEMV67326.2025.00030}
}

@inproceedings{rao1995bdi, 
    title={BDI Agents: From Theory to Practice},
    author={Rao, Anand S and Georgeff, Michael P and others},
    booktitle={Icmas},
    volume={95},
    pages={312--319},
    year={1995}
}

@article{laird1987soar, 
    title={SOAR: An architecture for general intelligence},
    author={Laird, John E and Newell, Allen and Rosenbloom, Paul S},
    journal={Artificial intelligence},
    volume={33},
    number={1},
    pages={1--64},
    year={1987},
    publisher={Elsevier},
    doi={10.1016/0004-3702(87)90050-6}
}

@inproceedings{kopp2006towards, 
    title={Towards a Common Framework for Multimodal Generation: The Behavior Markup Language}, 
    author={Kopp, Stefan and Krenn, Brigitte and Marsella, Stacy and Marshall, Andrew N and Pelachaud, Catherine and Pirker, Hannes and Th{\'o}risson, Kristinn R and Vilhj{\'a}lmsson, Hannes}, 
    booktitle={International workshop on intelligent virtual agents}, 
    pages={205--217}, 
    year={2006}, 
    organization={Springer}, doi={https://doi.org/10.1007/11821830_17}
}

@inproceedings{hartholt2013all, 
    title={All Together Now: Introducing the Virtual Human Toolkit}, 
    author={Hartholt, Arno and Traum, David and Marsella, Stacy C and Shapiro, Ari and Stratou, Giota and Leuski, Anton and Morency, Louis-Philippe and Gratch, Jonathan},
    booktitle={Int Workshop on Intelligent Virtual Agents}, 
    pages={368--381}, 
    year={2013}, 
    organization={Springer}
}

@inproceedings{park2023generative, 
    title={Generative Agents: Interactive Simulacra of Human Behavior}, 
    author={Park, Joon Sung and O'Brien, Joseph and Cai, Carrie Jun and Morris, Meredith Ringel and Liang, Percy and Bernstein, Michael S}, 
    booktitle={Proceedings of the 36th annual ACM symposium on user interface software and technology}, 
    pages={1--22}, 
    year={2023}
}

@misc{yao2022react,
  title={ReAct: Synergizing Reasoning and Acting in Language Models},
  author={Yao, Shunyu and Zhao, Jeffrey and Yu, Dian and Du, Nan and Shafran, Izhak and Narasimhan, Karthik and Cao, Yuan},
  year={2022},
  eprint={2210.03629},
  archivePrefix={arXiv},
  primaryClass={cs.CL},
  doi={10.48550/arXiv.2210.03629},
  url={https://arxiv.org/abs/2210.03629}
}

@incollection{nisiotis2023interwoven,
  title={Interwoven spaces with xr, ai, and robots: Merging realities in space and time},
  author={Nisiotis, Louis and Hadjiliasi, Aimilios and Alexandrou, Floris and Alboul, Lyuba},
  booktitle={Museums and Technologies of Presence},
  pages={243--261},
  year={2023},
  publisher={Routledge}
}

@article{yang2025embodied,
    title={Embodied Conversational Agents in Extended Reality: A Systematic Review},
    author={Yang, Fu-Chia and Acevedo, Pedro and Guo, Siqi and Choi, Minsoo and Mousas, Christos},
    journal={IEEE Access},
    year={2025},
    publisher={IEEE}, doi={10.1109/ACCESS.2025.3566698}
}

@article{
    baddeleyA1992WorkingMemory,
    author = {Alan Baddeley },
    title = {Working Memory},
    journal = {Science},
    volume = {255},
    number = {5044},
    pages = {556-559},
    year = {1992},
    doi = {10.1126/science.1736359},
    URL = {https://www.science.org/doi/abs/10.1126/science.1736359},
    eprint = {https://www.science.org/doi/pdf/10.1126/science.1736359},
}

@misc{tulving1972episodic,
  title={Episodic and Semantic Memory},
  author={Tulving, Endel},
  year={1972}
}

@misc{wang2025mirix,
  title={MIRIX: Multi-Agent Memory System for LLM-Based Agents},
  author={Wang, Yu and Chen, Xi},
  year={2025},
  eprint={2507.07957},
  archivePrefix={arXiv},
  doi={10.48550/arXiv.2507.07957},
  url={https://arxiv.org/abs/2507.07957}
}

@inproceedings{maharana2024evaluating,
  title={Evaluating Very Long-Term Conversational Memory of LLM Agents},
  author={Maharana, Adyasha and Lee, Dong-Ho and Tulyakov, Sergey and Bansal, Mohit and Barbieri, Francesco and Fang, Yuwei},
  booktitle={Proc. of the 62nd Annual Meeting of the Association for Computational Linguistics},
  pages={13851--13870},
  year={2024}
}

@misc{lu2024small,
  title={Small Language Models: Survey, Measurements, and Insights},
  author={Lu, Zhenyan and Li, Xiang and Cai, Dongqi and Yi, Rongjie and Liu, Fangming and Zhang, Xiwen and Lane, Nicholas D. and Xu, Mengwei},
  year={2024},
  eprint={2409.15790},
  archivePrefix={arXiv},
  doi={10.48550/arXiv.2409.15790},
  url={https://arxiv.org/abs/2409.15790}
}

@misc{belcak2025small,
  title = {Small Language Models are the Future of Agentic AI},
  author = {Belcak, Peter and Heinrich, Greg and Diao, Shizhe and Fu, Yonggan and Dong, Xin and Muralidharan, Saurav and Lin, Yingyan Celine and Molchanov, Pavlo},
  year = {2025},
  eprint = {2506.02153},
  archivePrefix= {arXiv},
  primaryClass = {cs.AI},
  url = {https://arxiv.org/abs/2506.02153},
  doi = {10.48550/arXiv.2506.02153}
}

@inproceedings{hadjiliasiCEAA2026,
  title={{CEAA}: A Cognitive Embodied Agents Architecture for Interactive Computing Systems},
  author={Hadjiliasi, Aimilios and Nisiotis, Louis},
  booktitle = {Proceedings of the 2026 IEEE 3rd International Symposium on Emerging Metaverse (ISEMV)},
  address   = {Cyprus},
  month     = oct,
  year      = {2026},
  note      = {In press}
}

@article{zhou2019edge,
    title={Edge Intelligence: Paving the Last Mile of Artificial Intelligence With Edge Computing},
    author={Zhou, Zhi and Chen, Xu and Li, En and Zeng, Liekang and Luo, Ke and Zhang, Junshan},
    journal={Proceedings of the IEEE},
    volume={107},
    number={8},
    pages={1738--1762},
    year={2019},
    publisher={IEEE},
    doi={10.1109/JPROC.2019.2918951}
}

@article{schick2023toolformer,
  title={Toolformer: language models can teach themselves to use tools},
  author={Schick, Timo and Dwivedi-Yu, Jane and Dess{\`\i}, Roberto and Raileanu, Roberta and Lomeli, Maria and Hambro, Eric and Zettlemoyer, Luke and Cancedda, Nicola and Scialom, Thomas},
  journal={Advances in neural information processing systems},
  volume={36},
  pages={68539--68551},
  year={2023}
}
\end{document}